\documentclass[10pt,twocolumn,letterpaper]{article}

\usepackage{cvpr}
\usepackage{times}
\usepackage{epsfig}
\usepackage{graphicx}
\usepackage{amsmath}
\usepackage{amssymb}
\usepackage{cvpr}
\usepackage{times}
\usepackage{epsfig}
\usepackage{graphicx}
\usepackage{amsmath}
\usepackage{amssymb}
\usepackage{times}
\usepackage{epsfig}
\usepackage{graphicx}
\usepackage{amsmath}
\usepackage{amssymb}
\usepackage{algorithm}
\usepackage{algorithmic}
\usepackage{multirow}
\usepackage{color}
\usepackage{color, colortbl}
\definecolor{Gray}{gray}{0.8}
\definecolor{Red}{rgb}{0.95,0.6,0.6}
\usepackage[breaklinks=true,bookmarks=false]{hyperref}

\cvprfinalcopy % *** Uncomment this line for the final submission

\def\cvprPaperID{****} % *** Enter the CVPR Paper ID here

\begin{document}

%%%%%%%%% TITLE
\title{DAP3D-Net: Where, What and How Actions Occur in Videos?}

\author{Li Liu \qquad Yi Zhou \qquad Ling Shao\\
Department of Computer Science and Digital Technologies\\
Northumbria University, Newcastle upon Tyne, NE1 8ST, UK\\
{\tt\small li2.liu@northumbria.ac.uk, m.y.yu@ieee.org, ling.shao@ieee.org}
}

\maketitle
%\thispagestyle{empty}

%%%%%%%%% ABSTRACT
\begin{abstract}
  Action parsing in videos with complex scenes is an interesting but challenging task in computer vision. In this paper, we propose a generic 3D convolutional neural network in a multi-task learning manner for effective \textbf{D}eep \textbf{A}ction \textbf{P}arsing (\textbf{DAP3D-Net}) in videos. Particularly, in the training phase, action localization, classification and attributes learning can be jointly optimized on our appearance-motion data via DAP3D-Net. For an upcoming test video, we can describe each individual action in the video simultaneously as: \textbf{Where} the action occurs, \textbf{What} the action is and \textbf{How} the action is performed. To well demonstrate the effectiveness of the proposed DAP3D-Net, we also contribute a new \textbf{N}umerous-category \textbf{A}ligned \textbf{S}ynthetic \textbf{A}ction dataset, i.e., \textbf{NASA}, which consists of $200,000$ action clips of more than 300 categories and with 33 pre-defined action attributes in two hierarchical levels (i.e., low-level attributes of basic body part movements and  high-level attributes related to action motion). We learn DAP3D-Net using the NASA dataset and then evaluate it on our collected \textbf{H}uman \textbf{A}ction \textbf{U}nderstanding (\textbf{HAU}) dataset. Experimental results show that our approach can accurately localize, categorize and describe multiple actions in realistic videos.
\end{abstract}

\begin{figure}
  \centering
  \includegraphics[width=0.455\textwidth,height=0.2625\textheight]{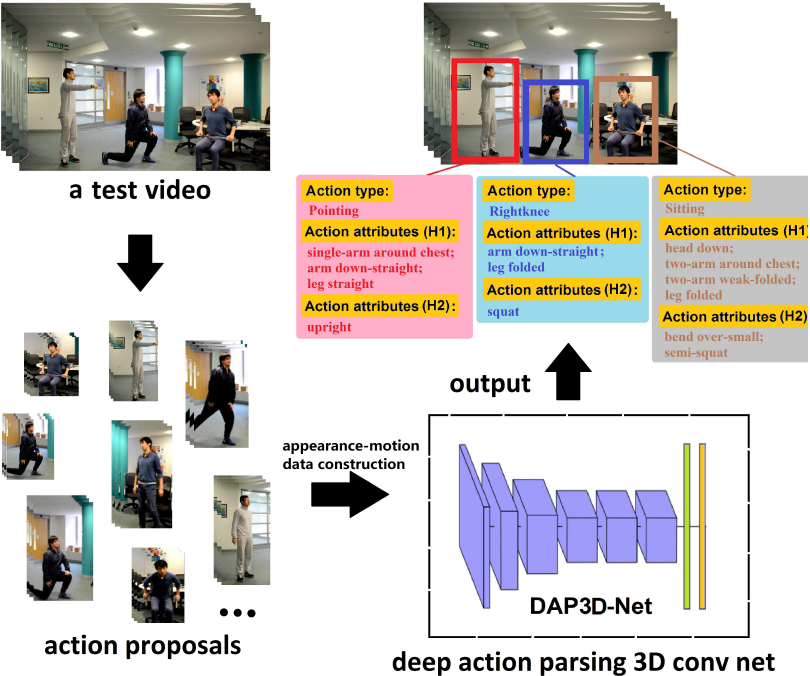}\\
  \caption{\footnotesize
\textbf{Diagram of the proposed work for action parsing}. Given a test video sequence, around 350 action proposals are extracted via \cite{yu2015fast}. Each proposal then goes through our proposed DAP3D-Net, which outputs these proposals' corresponding action types, action attributes (low-level (H1) and high-level (H2)) and refined action locations, simultaneously.}
  \label{intro}
  \vspace{-3.4ex}
\end{figure}

\vspace{-3ex}

%%%%%%%%% BODY TEXT
\section{Introduction}
 Human action analysis \cite{dollar2005behavior,xu2012detection,ni2015pose,fu2012attribute,antic2011video} is a popular research area in computer vision and has many applications such as video surveillance \cite{xulearning,zhang2008detection,ko2008survey}, robotics and multimedia search and retrieval \cite{yu2015fast,qin2015fast,yuan2011discriminative}. Detailed description of human actions in videos requires to solve three main problems: (1) \emph{Where in the video do the actions occur?} (2) \emph{What categories do the actions belong to?} and (3) \emph{How are these actions performed?}  Most of the previous studies, however, only focus on one or two of the problems separately (such as action categorization \cite{wang2011action,taylor2010convolutional,wang2014learning,liu2013learning,simonyan2014two,karpathy2014large,laptev2008learning}, localization \cite{klaser2012human,thi2010human,jain2014action,oneata2014spatio,weinzaepfel2015learning,van2015apt} or motion attributes learning \cite{liu2011recognizing,shao2015deeply}), and thus they cause poor generalization and high complexity to integrally describe actions with rich information in detail.

To jointly consider the above three problems, in this paper, we target to develop a new approach which can automatically parse actions in videos. Specifically, we are interested in describing each individual action in a video with its corresponding location, category and motion attributes, simultaneously as shown in Fig.~\ref{intro}. For localization of an individual action, we aim to output the accurate bounding box, in which an action occurs; for classification, we aim to categorize each action into a class; and for attributes learning, we describe how each action is performed with detailed motion information.  In fact, the three problems are inter-correlated and should be tackled together. However, little effort has been devoted to  action parsing in videos with complex scenes. In particular, there is no big enough aligned action data with corresponding motion attributes for model learning. Although some recent works \cite{zhang2013attribute,liu2011recognizing} annotated some action data with bounding boxes and attributes, the size and diversity of the data in their works are limited.

Thus, in this paper, we first contribute a new Numerous-category Aligned Synthetic Action dataset, i.e., NASA. It contains 200,000 action clips with over 300 categories. Moreover, each clip is assigned with 33 attributes in two hierarchical levels. All the video clips in NASA are synthesized using \emph{Poser10}\footnote{Details of \emph{Poser 10} in http://my.smithmicro.com/poser-10.html}, which is a professional software program used to effectively generate virtual action data from motion capture of real humans. Thus, the motions of synthetic data can be visually very close to realistic human actions. To the best of our knowledge, NASA is the largest aligned action dataset to date.

%\begin{figure}
%  \centering
%  \includegraphics[width=0.475\textwidth]{D11.pdf}\\
%  \caption{\footnotesize
%  \textbf{A glance of the NASA dataset}. Some example frames from the NASA action dataset are illustrated in left of the figure. All action categories ($300+$) are shown with tag cloud in right-top of the figure. Bigger font size of a category name indicates more data samples in this category. The defined semantic action attributes in NASA dataset are shown in the right-bottom as well. Red and blue represent the low-level attributes and high-level attributes, respectively.}
%  \label{nasa}
%  \vspace{-4ex}
%\end{figure}

From the perspective of modeling our action parsing problem, we have been motivated by region-based convolutional neural network (R-CNN) \cite{girshick2014rich,girshick2015fast,ren2015faster} for object detection problems, since deeply learned models have proved to achieve better results than conventional methods \cite{wang2013regionlets,uijlings2013selective,fidler2013bottom,felzenszwalb2008discriminatively}. Furthermore, in \cite{shao2015deeply,shankar2015deep}, learning attributes from complex scene images via deep nets can also produce significant improvement over traditional methods. However, most of current deep learning based approaches focus on 2D images. For spatio-temporal action data, some previous works \cite{taylor2010convolutional,tran2014c3d,ji20133d,karpathy2014large} have proposed 3D deep convolutional neural networks to recognize actions. Among them, \cite{tran2014c3d} produces better results with 3D convolutions on both spatial and temporal dimensions. Inspired by all these works, we aim to design a multi-task 3D convolutional neural network for effective \textbf{D}eep \textbf{A}ction \textbf{P}arsing (\textbf{DAP3D-Net}) in videos. Specifically, in the training phase, action localization, classification and attributes learning can be jointly optimized via DAP3D-Net. Once model training is completed, given an upcoming test video, we can describe each individual action in the video simultaneously as: where the action occurs, what the action is and how the action is performed. Different from \cite{tran2014c3d,ji20133d,taylor2010convolutional} using original video data as the deep net inputs, to better learn motion information for action parsing, in our method, we adopt motion channels (calculated from optical flows) in addition to appearance (intensity information) as the input of DAP3D-Net. The promising action parsing results achieved by DAP3D-Net provide potential functionalities of outdoor/indoor video surveillance systems for public security and personal healthcare applications, e.g., anomalous event detection and abnormal behavior monitoring for elderly. The main contributions of our work can be highlighted as follows:

(1) Our deep model DAP3D-Net jointly optimizes action localization, classification and attributes learning. With such a multi-task scheme, DAP3D-Net can solve the problem of where, what and how actions occur simultaneously for human action parsing in videos.

(2) In order to train  DAP3D-Net, we introduce a new large-scale aligned action dataset, NASA, with  200K well labeled video clips. Additionally, to further evaluate the effectiveness of DAP3D-Net, a realistic Human Action Understanding (HAU) dataset has also been collected with the locations, categories and attributes of all actions annotated.

% \vspace{-1ex}
%\begin{itemize}
%  \item Our deep model DAP3D-Net can jointly optimize action localization, classification and attributes learning. With such a multi-task scheme, DAP3D-Net can solve the problem of where, what and how actions occur simultaneously for human action parsing in videos.
%    \vspace{-1ex}
%  \item In order to train  DAP3D-Net, we introduce a new large-scale aligned action dataset, NASA, with  200K well labeled video clips. Additionally, to further evaluate the effectiveness of DAP3D-Net, a realistic Human Action Understanding (HAU) dataset has also been collected with the locations, categories and attributes of all actions annotated.
%\end{itemize}
%\vspace{-1ex}

\section{Related Work}
Since little work has been done on action parsing in videos for simultaneously solving the problems of action localization, categorization and attributes learning, in this section, we mainly review some related work on action detection and action attributes modeling. Action detection can be regarded as a combination of action localization and categorization. In \cite{siva2011weakly}, a weakly supervised model with multiple instance learning was applied for action detection. In \cite{wang2014video}, a dynamic-poselets method was introduced. Branch-and-bound algorithm \cite{yuan2011discriminative} was proposed to reduce the action detection complexity. There also exist some sub-volume \cite{lan2011discriminative,ke2007event,siva2010action} based action detection methods. Besides, the cross-dataset action detection \cite{cao2010cross} and the spatio-temporal deformable part models based action detection \cite{tian2013spatiotemporal} have also been proposed in previous studies. Additionally, action detection via fast proposals was developed in \cite{yu2015fast}.

For action attributes modeling, Liu et al. \cite{liu2011recognizing} used high-level semantic attributes to represent human actions in videos and further constructed more descriptive models for the action recognition task. A similar idea has also been applied in \cite{zhanghuman,tahmoush2015applying} for improved action categorization. Moreover, a convolutional multi-task learning method \cite{zhang2013attribute} has been adopted for action recognition from low-level features with attribute regularization. In \cite{zhang2013robust}, a robust learning framework using relative attributes was developed for human action recognition. Additionally, action attributes and object-parts from images were also used for action recognition in \cite{yao2011human}. However, all the above studies mainly focus on action recognition by means of attributes rather than general action attributes learning tasks. Although, in \cite{fu2012attribute}, authors have jointly tackled the  classification and attributes annotation for group activities, it is still regarded as a separated feature extraction and attribute learning pipeline rather than an end-to-end framework as our multi-task DAP3D-Net. Besides, DAP3D-Net focuses on simultaneous localization and parsing of multiple actions, while in \cite{fu2012attribute} only global representations of group activities are considered.

\section{Approach}
In this section, we introduce the architecture of our proposed method, i.e., DAP3D-Net, for action parsing in videos. It is regarded as a multi-task learning scheme for jointly optimizing action localization, action categorization and action attributes learning using appearance-motion data from the NASA dataset. In the following subsections, we will first describe the construction of the NASA d ataset and then detail the DAP3D-Net.

\subsection{NASA Dataset Construction}
For deep action parsing in videos, a large  number of aligned action clips with motion attributes are needed to train the models. However, most of the previous action datasets \cite{liu2009recognizing,marszalek2009actions,kuehne2011hmdb,soomro2012ucf101,karpathy2014large} are collected with realistic scenarios, in which complex background, camera noise, shift, scaling and occlusion always exist. Such action videos  indeed benefit for evaluating the robustness of action recognition systems, but are not suitable as the training data to learn a model for action parsing, since they lack well localized bounding boxes and annotations for each individual action in the videos.  In fact, most of previous action datasets are collected from websites with realistic scenes of activities rather than just aligned single actions. For instance, the largest action dataset so far, Sport1M \cite{karpathy2014large}, is collected from YouTube, which contains one million clips from 487 sport activity categories, but with no particular annotation for each individual action due to the fact that detailed annotation for large-scale video data is laborious and impractical. The similar circumstances also exist in other realistic action datasets \cite{soomro2012ucf101,marszalek2009actions}. On the contrary, there are indeed some earlier action datasets \cite{schuldt2004recognizing,blank2005actions,yuan2011discriminative}  with the bounding boxes given or easily obtainable on relatively simple backgrounds, e.g., the KTH dataset (with 6 basic action categories), however, the size and diversity of these action datasets are too limited to train a sophisticated action parsing model.

To solve the shortage of aligned and annotated action data for training models with some specific usage, e.g., action parsing proposed in this paper, we contribute a new dataset, i.e., NASA, which provides 200,000 aligned action clips from over 300 action categories. For each clip, 33 action attributes are assigned in two hierarchical levels. The NASA dataset is superior to previous aligned action datasets in both scale and diversity.

\vspace{-0.9ex}
\subsubsection{Action Synthesis}
\vspace{-0.9ex}
As mentioned above, it is impractical to construct a large-scale aligned action dataset by either annotating actions from online data or newly recording action videos by actors. Naturally, computer graphics and animation techniques can be used  to synthesize  human actions  \cite{ragheb2008vihasi}  automatically. Therefore, we adopt \emph{Poser 10}, which is a professional software for creating 3D animation and illustration. With \emph{Poser 10}, we can build high quality character-based 3D action animations and further render the animations into photo-realistic videos or images. Moreover, a truebones motion capture database\footnote{Download from http://www.truebones.com} is also utilized for \emph{Poser 10}, in which over 1600 different high-quality, clean and consistent motion models captured from realistic human performers are packed with BVH files. Specifically, a 3D model in a BVH file keeps movement parameters with 19 joints which are captured from 19 corresponding  body parts of a real human with wearable devices. We further aggregated 1600 motion models into over 300 action categories and then imported these models into \emph{Poser 10} and generated the action clips, in each of which only one aligned action is included. In particular, for each 3D motion model, we projected it into 18 viewpoints with three different camera-heights and, in each camera-height, circle-distributed six  viewpoints were captured from 3D action models. To increase the diversity of actions, \emph{Poser 10} also  provides ten different characters with different genders, looks, body sizes, heights and clothing to perform the actions in various backgrounds. In this way, for a certain 3D action model from truebones motion capture database, we can create $3 (camera heights)\times6(view points)\times10 (characters)=180$ action clips. The spatial size of each clip is fixed as $342\times256$ via \emph{Poser 10}, while video lengths are assigned based on action categories. Since all action motion models are captured from real human body movements, the synthetic clips in NASA are visually very close to realistic actions performed by humans.

\begin{figure}
  \centering
  \includegraphics[width=0.47\textwidth,height=0.2\textheight]{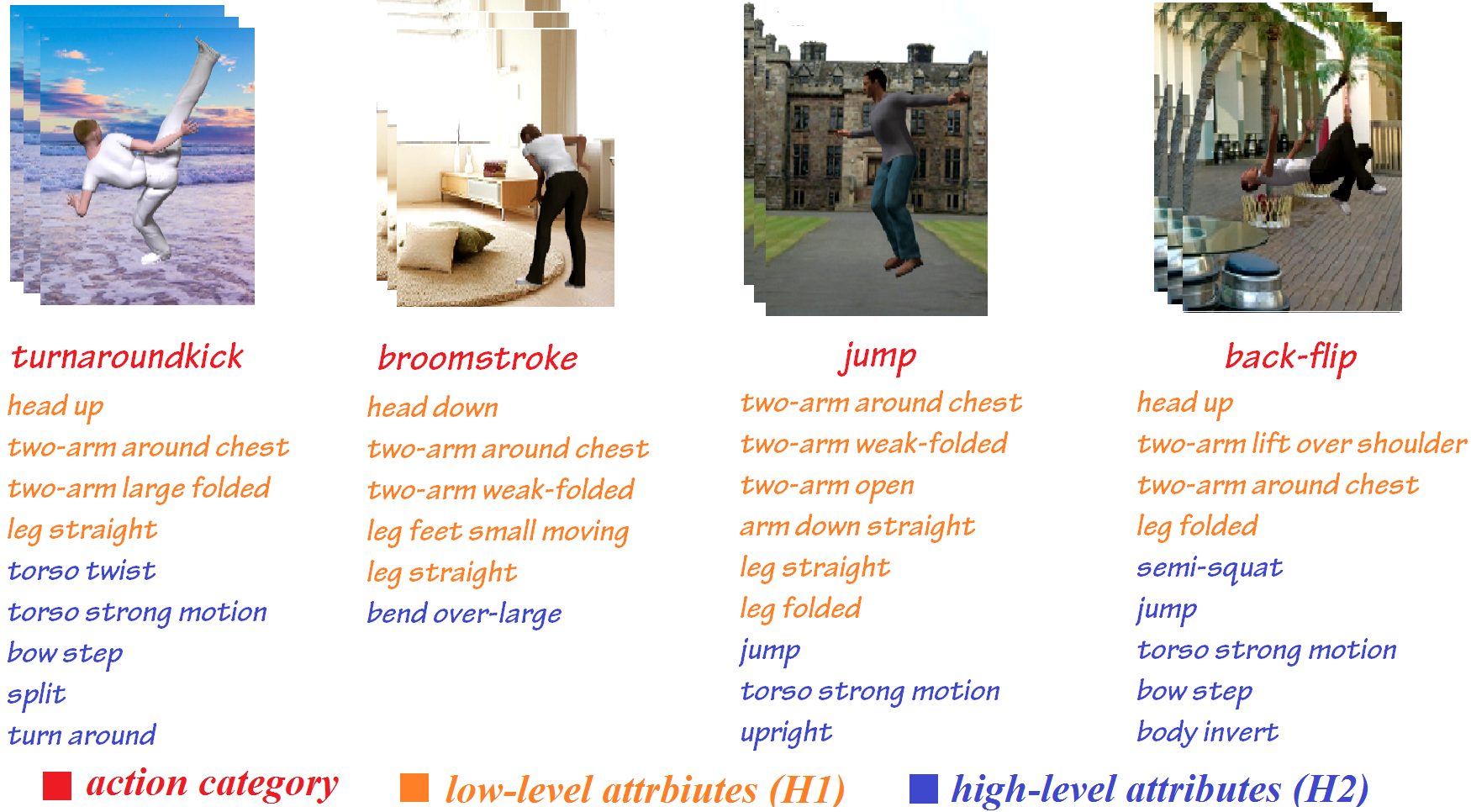}\\
  \caption{\footnotesize
  \textbf{Video examples in our NASA dataset}. Each of the labeled video is assigned with multiple semantic action attributes in hierarchical levels: low-level (H1) and high-level (H2).}
  \label{example}
  \vspace{-3ex}
\end{figure}

\vspace{-1ex}

\subsubsection{Action Attribute Annotation}
\vspace{-0.5ex}
After data generation by \emph{Poser 10}, we further annotate each clip with 33 action attributes in two hierarchical levels. Specifically, we define 19 low-level attributes (H1) to describe the basic body part movements and 14 high-level attributes (H2) of general action motion.
%Low-level: \emph{head up, head down, head turning, single arm lift over shoulder, double arms lift over shoulder, single arm around chest, double arms around chest, single arm large-fold,double arms large-fold, single arm weak-fold, double arms weak-fold, arm down-straight, arm swing-move back-forward motion (straight), arm swing-move back-forward motion (fold), arm open, leg pendulum-like motion (kick), leg feet small moving, leg straight, leg fold};
%
%High-level: \emph{large-bend over, weak-bend over, semi-squat, squat, jumping, torso twist, torso strong motion, bow step, splits, body invert, hip twist, turn round, sit/lie and hand touch the ground, upright-based}.
%\end{small}
Since the action clips generated from each truebones model are motion consistent, we regard these 33 attributes as the \emph{model-level attributes} in our NASA dataset. In this way, all the data generated from one model will \emph{share} the same 33 action attributes\footnote{Since each category in NASA is aggregated from multiple models, all attributes for a category will be diverse rather than sharing the same ones.}. However, in some cases, the attributes of action clips are not dominant enough for their corresponding action categories due to the downloaded models with inaccurate motion capture. Thus, to control the data quality, we discard clips with ambiguous attributes and finally construct our NASA dataset with  200,000 action clips. Given a  video clip, the semantic motion attributes can allow us to well parse an action by answering ``\emph{How are these actions performed?}" and also may benefit zero-shot learning \cite{jayaraman2014zero,lampert2014attribute} for action recognition/retrieval in future work. Fig.~\ref{example} shows some video examples with two level attributes in the NASA dataset. The full list of each action clip's attributes will be released in our project webpage later.

\subsection{Modeling DAP3D-Net}
In this sub-section, we introduce our proposed deep architecture, i.e., DAP3D-Net, for effective action parsing by solving the problems of ``Where, What and How Actions Occur in Videos?". In particular, the learning phase of the proposed approach consists of three steps: (1) Data augmentation via video spatial cropping and temporal scaling; (2) Appearance-motion data composition;  (3) A 3D convolutional neural network for joint optimization of action localization, classification and attributes learning.

\vspace{-2ex}

\subsubsection{Preliminary work for training data}\label{rel}
\textbf{Data augmentation:} To better resist overfitting and train a effective bounding box regressor (which will be explained in Section.~\ref{rel2}) with DAP3D-Net for action localization, we crop the each aligned action clip in NASA into 5 subclips. Specifically, given a clip with $S=(d_{w},d_{h},d_{l})$, where $d_{w}$, $d_{h}$ and $d_{l}$ indicate the width, height and temporal length of the clip, respectively, we crop it spatially into five subclips, i.e., top-left ($s1$), top-right ($s2$), bottom-left ($s3$), bottom-right ($s4$) and central ($s5$), with the identical width of $d_{w}-2t$ and height of $d_{h}-2t$. In detail,  we define the spatial center coordinate of an original clip as $(0,0)$ and according to this center position, the relative location of each subclip $(d_{w\_shift},d_{h\_shift})$ can be denoted as: $(-t,-t)$ for $s1$, $(t,-t)$ for $s2$, $(-t,t)$ for $s3$, $(t,t)$ for $s4$ and $(0,0)$ for $s5$. Since each action is well aligned in data, the relative location $(d_{w\_shift},d_{h\_shift})$ of subclips can be regarded as the lengths of spatial shifts along horizonal and vertical directions respectively from the central position $(0,0)$ of original clips in NASA. In our model, $d_{w}=256$, $d_{h}=342$, and for each clip in NASA we crop it into subclips by using a parameter $t$ randomly selected with $t\in(10,50)$. To further achieve temporally scale invariant action parsing via DAP3D-Net, we also resize these subclips along the temporal dimension with $d_{l}\rightarrow pd_{l}$, where $p$ is a scaling factor randomly selected from $p\in(0.5,1.5)$ for each subclip.

\begin{figure}
  \centering
  \includegraphics[width=0.43\textwidth]{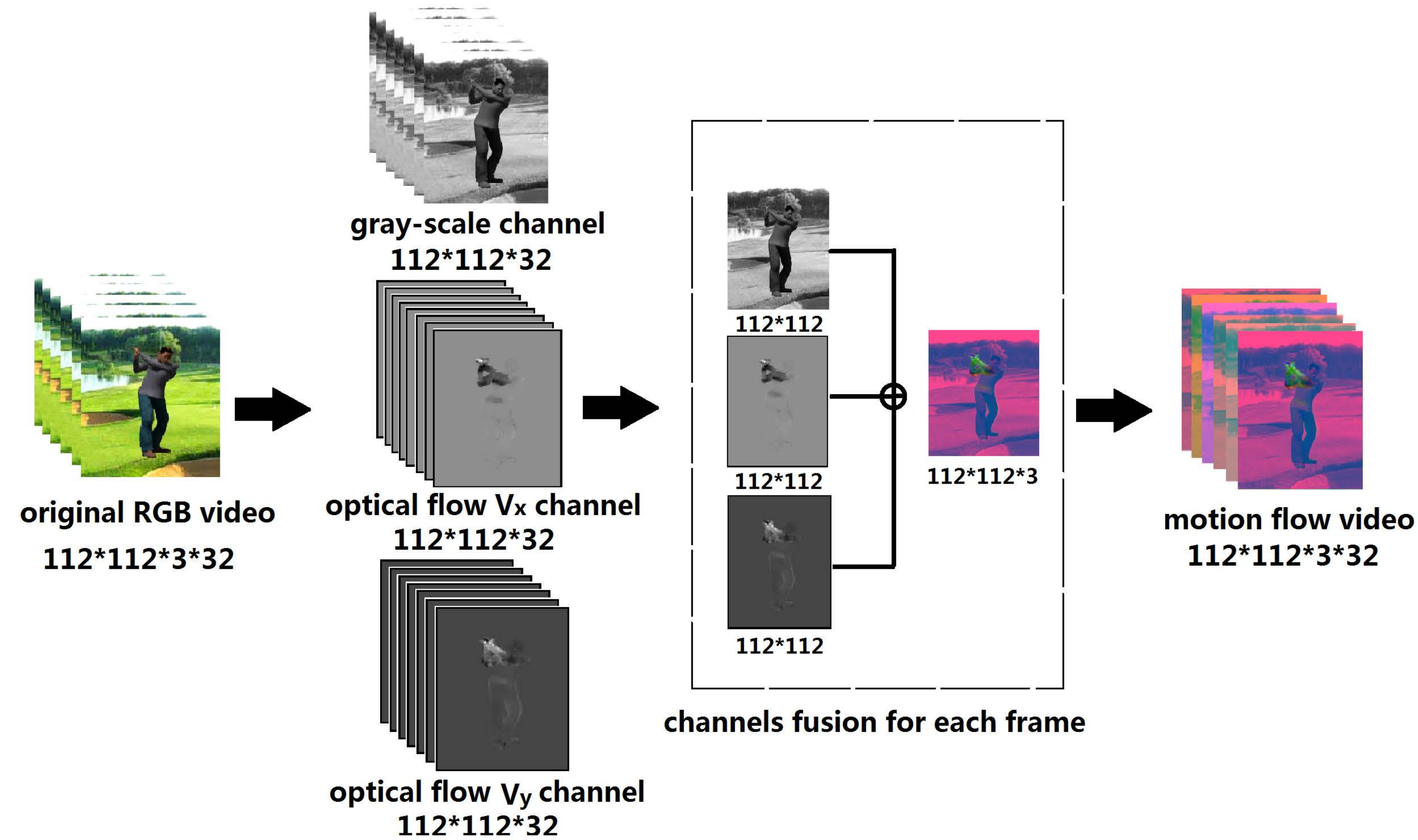}\\
  \caption{\footnotesize
  \textbf{Illustration of appearance-motion data composition.}}
  \label{SM}
  \vspace{-3.5ex}
\end{figure}

\textbf{Appearance-motion data composition:} Since our work focuses on action parsing rather than group activity recognition, temporal motion information appears to be more important than spatial appearance information from one action clip. Therefore, after data augmentation, we propose to construct appearance-motion data instead of using the traditional RGB videos as inputs to our DAP3D-Net. Specifically, we first extract the intensity information via gray-scale transformation as the appearance channel, and further compute the optical flows ($V_{x}$ and $V_{y}$) via \cite{liu2009beyond}, which carry rich motion information from videos, as the motion channels. We then combine one intensity channel with two motion channels to construct our appearance-motion  video data for DAP3D-Net training as shown in Fig.~\ref{SM}. In fact, appearance-motion data can achieve the theoretical similar effects as ``early fusion" \cite{karpathy2014large} of RGB and motion data in other deep models, and more importantly reduce the artifacts caused by synthetic data to the greatest extent, since in this composition appearance information is weakened but motion information is strengthened.

\begin{figure*}
  \centering
  \includegraphics[width=1\textwidth,height=0.18\textheight]{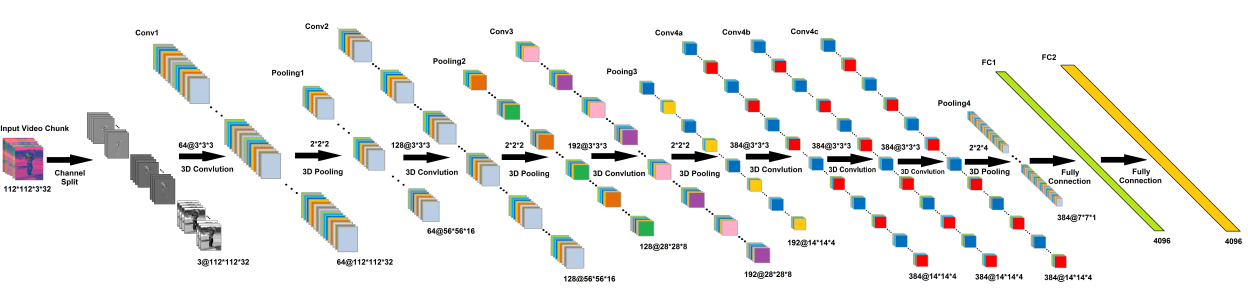}\\
  \caption{\footnotesize
  \textbf{DAP3D-Net structure}. The rectified linear unit (RELU) is applied as active function after each convolution layer and fully connected layer. The detailed parameters of DAP3D-Net can be found in this figure. (Zoom in for better viewing.) }
  \label{net}
    \vspace{-2.5ex}
\end{figure*}

\vspace{-1.5ex}

\subsubsection{DAP3D-Net Structure and Model Setting}\label{rel2}
%Our proposed deep net structure is illustrated as Fig.~\ref{net}. With feasible GPU memory, we design our DAP3D-Net with
%have 6 convolution layers \emph{Conv(N,K,S)}: with \emph{N} feature maps, kernel size \emph{K} and stride \emph{S}; 4 pooling layers \emph{Pooling(N,K,S)}: with \emph{N} feature maps, kernel size \emph{K} and stride \emph{S}, followed by 2 fully connected layers \emph{FC(N)}: with \emph{N} elements.  All 3D convolution kernels are performed with stride 1 along both spatial and temporal dimensions. The rectified linear unit (RELU) is applied as active function after each convolution layer. We conclude the DAP3D-Net with the parameters as: \emph{Conv(64,3,1)}-\emph{Pooling(64,2,1)}-\emph{Conv(128,3,1)}-\emph{Pooling(128,2,1)}-\emph{Conv(192,3,1)}-\emph{Pooling(192,2,1)}-\emph{Conv(384,3,1)}-\emph{Conv(384,3,1)}-\emph{Conv(384,3,1)}-\emph{Pooling(384,2,1)}-\emph{FC(4096)}-\emph{FC(4096)}.

\begin{figure}
  \centering
  \includegraphics[width=0.4\textwidth,height=0.165\textheight]{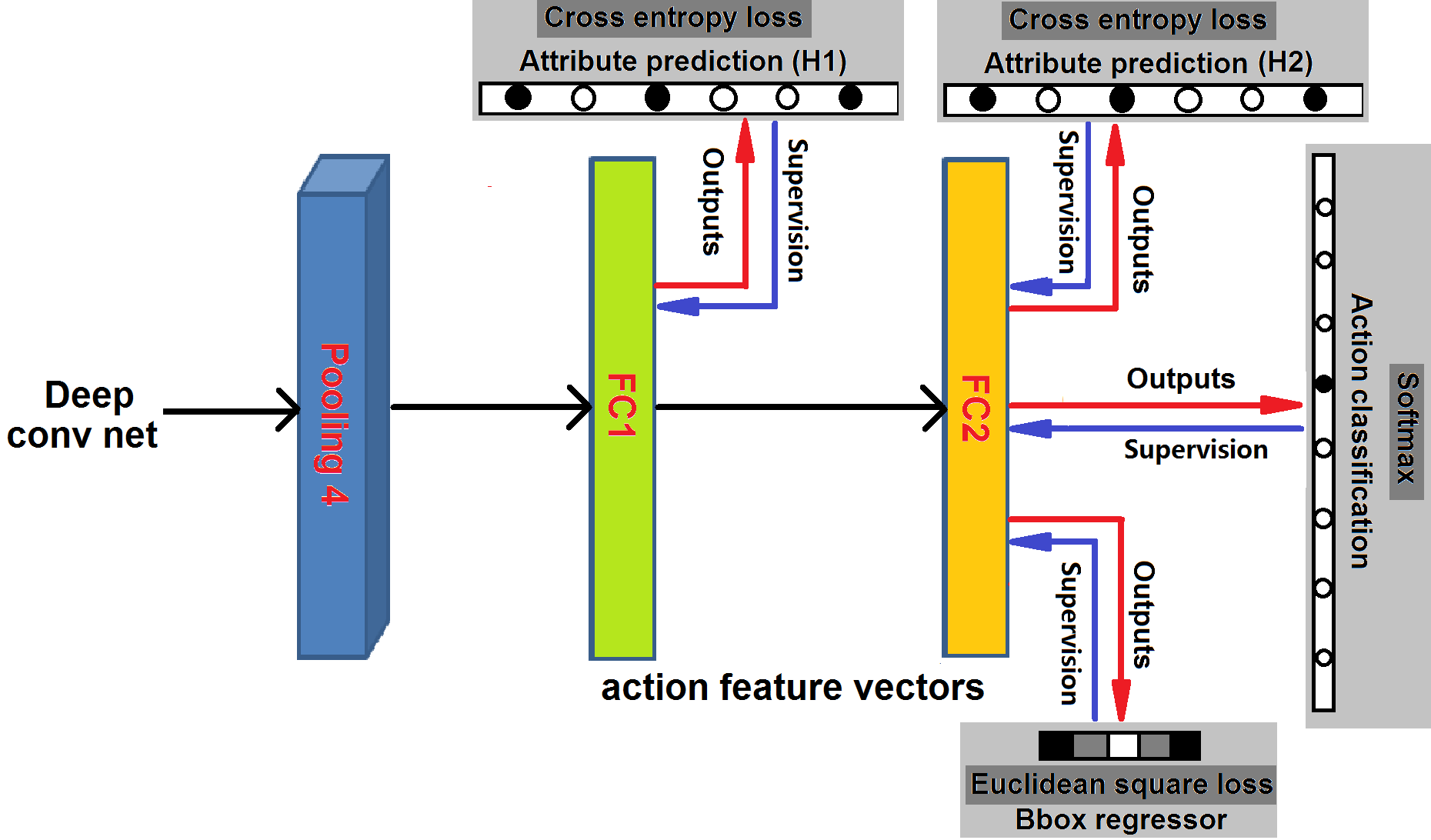}\\
  \caption{\footnotesize
  \textbf{The multi-task loss supervision (output) of DAP3D-Net}.}
  \label{M-loss}
  \vspace{-3ex}
\end{figure}

The proposed structure of DAP3D-Net contains 6 (spatio-temporal) convolution layers, 4 pooling layers and 2 fully connected (FC) layers. All the parameters are illustrated in detail in Fig.~\ref{net}. As shown in Fig.~\ref{net}, an input appearance-motion sequence is fed to DAP3D-Net and mapped into feature vectors by fully connected layers (FC1 and FC2). After FC layers, we define four multi-task loss supervision (output) layers in DAP3D-Net for our action parsing task as shown in Fig.~\ref{M-loss}. In more detail, a two-dimensional relative location vector, i.e., $loc=(\widehat{d}_{w\_shift},\widehat{d}_{h\_shift})$, as mentioned in Section~\ref{rel}, is optimized via Euclidean square loss  for real-valued bounding-box regression connected with FC2. Furthermore,  a softmax loss layer is applied to predict probabilities, $p^{c}=\{p^{c}_{0}\ldots p^{c}_{M-1}\}$, of $M$ action categories connected with FC2. Besides, following \cite{shao2015deeply}, the cross entropy loss is optimized for multi-output prediction of 14 high-level action attributes (H2), $p^{h2}=\{p^{h2}_{0}\ldots p^{h2}_{13}\}$, connected with FC2, and similarly 19 low-level action attributes prediction (H1), $p^{h1}=\{p^{h1}_{0}\ldots p^{h1}_{18}\}$, is connected with FC1.  Particularly, here we also explain the reasons why we split attributes into two layers as follows: this is because FC1 is relatively closer to the low-level feature extraction layers in DAP3D-Net (i.e., \emph{Conv1},\emph{Conv2} and \emph{Conv3}) and involves weak semantic information compared with FC2, thus FC1 is more suitable for learning attributes of basic body parts' movements which are relatively lower-level and more specific. On the contrary, since FC2 is regarded as a combination of low-level information from FC1 and directly connected with the action category prediction layer, it contains more high-level, semantic and abstract information of actions. Thus, we can learn higher-level, general motion attributes from FC2. This hierarchical learning scheme proves to be effective and accurate for revealing the motion details of actions in our experiments.

\vspace{0.125ex}

For training DAP3D-Net, each input subclip is labeled with a relative location, a ground-truth class label and multiple attributes in H1 and H2. Then, we adopt a multi-task loss $L_{DAP3D}$ on each training datum to jointly optimize bounding-box regression, action categorization and attributes learning as follows:
%\begin{equation}
%\small
%\begin{split}
%\label{eq:e1}
%&L_{All}(loc,p^{c},p^{h1},p^{h2})=\\
%&L_{cat}(p^{c})+\lambda_{1}L_{attr(H1)}(p^{h1})+\lambda_{2}L_{attr(H2)}(p^{h2})+\beta L_{bbox}(loc)
%\end{split}
%\end{equation}
\begin{small}
\begin{align}
\label{eq:e1}
&L_{DAP3D}(loc,p^{c},p^{h1},p^{h2})= \\
&L_{cat}(p^{c})+\lambda_{1}L_{attr(H1)}(p^{h1})+\lambda_{2}L_{attr(H2)}(p^{h2})+\beta L_{bbox}(loc) \nonumber
\end{align}
\vspace{-4ex}
\end{small}

where, $L_{cat}(p^{c})$ indicates the softmax loss of action categorization which is defined the same as other deep classification nets \cite{krizhevsky2012imagenet,karpathy2014large,szegedy2014going}.
$L_{attr(H1)}$ indicates the cross entropy loss for hierarchical attributes learning as:
\vspace{-1ex}
\begin{equation}
\small
L_{attr(H1)}=-\frac{1}{N}\sum_{i}^{N}t^{h1}_{i}\log p^{h1}_{i}+(1-t^{h1}_{i})\log (1-p^{h1}_{i})
\end{equation}
where $N=19$ denotes the number of the H1 attribute outputs, $t^{h1}_{i} (i=1\ldots N)$ are ground-truth labels and $p^{h1}_{i} (i=1\ldots N)$ are output probability predictions. A similar equation is also used for $L_{attr(H2)}$. Furthermore, following \cite{girshick2014rich}, the Euclidean square loss $L_{bbox}$ for bounding box regression can be defined as:
\vspace{-1ex}
\begin{equation}
\small
L_{bbox}=||(loc-loc_t)||^{2}
\vspace{-1ex}
\end{equation}
where $loc=(\widehat{d}_{w\_shift},\widehat{d}_{h\_shift})$ is the predicted relative location and $loc_t=(d_{w\_shift},d_{h\_shift})$ is the ground-truth relative location. Besides, the hyper-parameters $\{\lambda_{1}, \lambda_{2}, \beta\}$ in Eq. (\ref{eq:e1}) are used to control the balance of multi-task losses. They are fixed as: $\lambda_{1}=0.5$, $\lambda_{2}=0.5$ and $\beta=0.5$ in all our experiments. Note that, to fit DAP3D-Net, we adopt affine spatio-temporal warping (similar as in \cite{girshick2014rich}) to resize all appearance-motion clips with the  fixed-size $112\times112\times3\times32$ as the inputs\footnote{In \cite{tran2014c3d}, they split videos into 16-frame long clips with a 8-frame overlap between two consecutive clips as the inputs of the deep net. The reason is \cite{tran2014c3d} focuses on group activity recognition and the information of the activity scene on spatial dimensions of a clip is more important than information on the temporal dimension. Thus, splitting with 16-frames does not break the integrity of target activities in training data. However, each training video in NASA includes only one integrated action and any split will make it incomplete. Thus, we warp all training clips with the same size as the DAP3D-Net inputs instead of brutally splitting them.} to the deep model. After attempting various network architectures with different parameter settings, the current 3D convolutional neural network structure shown in Fig.~\ref{net} proves to be the best option for our action parsing task.
%\begin{equation}
%\small
%L_{bbox}(loc)=f_{L1}(t_{w-shift}-d_{w-shift})+f_{L1}(t_{h-shift}-d_{h-shift})
%\end{equation}
%in which
%
%
%\begin{equation}\label{plabel}
%\small
%  f_{L1}(x) = \left\{\begin{array}{l}
%                       0.5x^{2},~  \text{if} |x|<1 \\
%                       |x|-0.5,~  \text{otherwise}
%                     \end{array}
%  \right.,
%\end{equation}

%\begin{figure*}
%  \centering
%  % Requires \usepackage{graphicx}
%  \begin{tabular}{cc}
%    \includegraphics[width=0.49\textwidth]{51.pdf} &\includegraphics[width=0.49\textwidth]{52.pdf}\\
%    \footnotesize
%    (a) Action categorization via DAP3D-Net$^{1}$ (RGB data)& \footnotesize (b) Action categorization via DAP3D-Net$^{2}$ (appearance-motion data)\\
%  \end{tabular}
%  \caption{\footnotesize
%  \textbf{Accuracy (\%) of each action category via DAP3D-Net$^{1}$ and DAP3D-Net$^{2}$, respectively}. (Zoom in for better viewing.)}
%  \label{nasaacc}
%  \vspace{-2ex}
%\end{figure*}

\begin{figure*}
  \centering
  \includegraphics[width=1\textwidth,height=0.126\textheight]{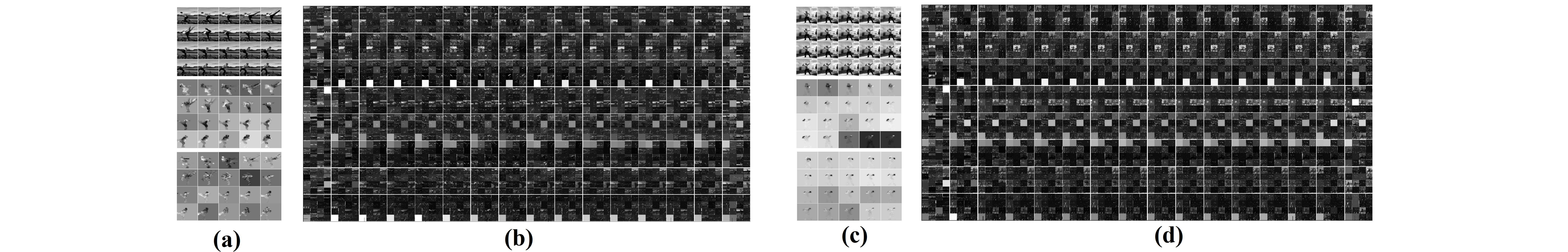}\\
  \caption{\footnotesize
  \textbf{Visualization of the feature map in \emph{Conv2} layer.} (a) The three channels (\emph{from top to bottom -} gray-scale, optical flow Vx and optical flow Vy) of 20-frame appearance-motion sequence: \emph{Turnaroundkick}; (b) the $128@56\times56\times16$ feature map of \emph{Conv2} layer for action \emph{Turnaroundkick}. In particular, \emph{Conv2} feature maps are illustrated in a total of $8\times16=128$ blocks and each block is the visualization of a $56\times56\times16$ dimensional feature. The similar visualization of feature maps for a appearance-motion sequence: \emph{bowandarrow} in \emph{Conv2} layer can also be seen in (c) and (d). (Zoom in for better viewing.)}
  \label{featuremap}
  \vspace{-2ex}
\end{figure*}

\subsection{Action Parsing via DAP3D-Net}
Once DAP3D-Net is trained, for a new test video containing multiple actions with complex scene, we first apply an action proposal method to obtain a set of candidate action chunks with corresponding coarse locations motivated by \cite{girshick2014rich}. In fact, a variety of recent papers offer methods \cite{jain2014action,oneata2014spatio,weinzaepfel2015learning,yu2015fast,van2015apt} for generating action proposals to effectively reduce the searching space for detection-based research. In our framework, we adopt \cite{yu2015fast} to propose the action candidates (average 550 proposals per video) due to the impressive performance reported in their paper.  We further construct the appearance-motion data and warp them into the identical size of $112\times112\times3\times32$ for each obtained action proposal as in the training phase, and then feed them into the learned DAP3D-Net. Consequently, the relative location, action category, and action attributes of each proposed video chunk can be obtained through DAP3D-Net. Particularly, to further refine the localization of actions, we formulate the bounding box adjustment equation as:
\vspace{-0.75ex}
\begin{equation}
\label{eq:e2}
\small
Loc_{ref}=(\widehat{d}_{w\_shift}+X_{pro},\widehat{d}_{h\_shift}+Y_{pro},T_{pro})
\vspace{-0.75ex}
\end{equation}
where $Loc_{ref}$ denotes the refined location of the action center,  coordinate $(X_{pro},Y_{pro},T_{pro})$ denotes the center location of the action proposal and coordinate $(\widehat{d}_{w\_shift},\widehat{d}_{h\_shift})$ denotes the relative location outputted from DAP3D-Net. Finally,  a non-maximum suppression is applied to reject a proposal if it has an intersection-over-union (IoU) overlap with a higher scoring proposal and the IoU is larger than a threshold following \cite{girshick2014rich}. For a faster action parsing, SVD can be applied with FC layers as \cite{girshick2015fast}.

\begin{table*}
\center
\newcommand{\tabincell}[2]{\begin{tabular}{@{}#1@{}}#2\end{tabular}}
\caption{\footnotesize
\textbf{Detection average precision (\%) on the HAU dataset}. Four previous works and some baselines have been evaluated against in this table. All the methods from the 5th row to the last row are based on action proposals in \cite{yu2015fast}. The last four rows are different versions of our proposed method.}
\label{table:t4}
\resizebox{0.98\textwidth}{0.071\textheight}{
\begin{tabular}{|c|cccccccccccccccccccc|>{\columncolor{Red}}c|}
\hline
\textbf{Method} &\rotatebox{0}{\textbf{\tabincell{c}{baseball\\pitch}}} &\rotatebox{0}{\textbf{\tabincell{c}{basketball\\control}}} &\rotatebox{0}{\textbf{\tabincell{c}{basketball\\shooting} }}&\rotatebox{0}{\textbf{boilrstok}} &\rotatebox{0}{\textbf{\tabincell{c}{bowand\\arrow}}} &\rotatebox{0}{\textbf{clapping }}&\rotatebox{0}{\textbf{\tabincell{c}{crouch\\walk} }}&\rotatebox{0}{\textbf{\tabincell{c}{golf\\swing}} }&\rotatebox{0}{\textbf{jump}} &\rotatebox{0}{\textbf{\tabincell{c}{karate\\kick}}} &\rotatebox{0}{\textbf{pointing} } &\rotatebox{0}{\textbf{\tabincell{c}{pull\\rope}}} &\rotatebox{0}{\textbf{\tabincell{c}{right\\knee}}}&\rotatebox{0}{\textbf{rounhouse }}& \rotatebox{0}{\textbf{\tabincell{c}{scurry\\jog}}} & \rotatebox{0}{\textbf{sitting}}&\rotatebox{0}{\textbf{skiing}}&\rotatebox{0}{\textbf{soccer}}&\rotatebox{0}{\textbf{surfing}} &\rotatebox{0}{\textbf{walk}} &\textbf{MAP}\\
 \hline
 \hline
 Tubelet \cite{jain2014action} &9.8&54.0&40.4&88.3&71.6&52.7&50.0&61.7&15.8&82.8&43.5&57.1&63.4&18.5&19.4&64.1&\textbf{94.8}&82.3&65.5&75.8&55.6\\
 SDPM \cite{tian2013spatiotemporal}&20.2&52.9&21.3&75.0&54.4&47.2&43.1&58.2&14.9&61.1&49.3&78.1&56.1&38.5&21.8&60.2&87.6&57.3&46.8&62.8&50.3\\
 WSAD \cite{siva2011weakly}&8.2&43.7&19.3&63.8&57.6&34.7&36.9&68.8&14.0&62.7&31.6&49.5&52.6&7.9&16.4&41.7&62.1&65.5&36.0&64.2&41.8\\
NBMIM \cite{yuan2011discriminative}&11.7&58.8&47.6&86.9&77.8&48.6&56.5&56.7&10.4&82.1&37.7&43.3&62.4&8.6&26.7&58.0&82.0&\textbf{79.7}&62.4&73.9&53.5\\
 \hline
  \hline
 \tabincell{c}{STIP \cite{laptev2008learning}+VLAD \cite{jegou2010aggregating} + linear SVM}&27.0&45.9&12.2&38.9&52.0&57.0&42.1&23.1&42.5&42.1&56.1&52.9&20.0&24.8&50.9&29.9&23.8&48.5&14.8&33.2&37.0\\
 \tabincell{c}{DTF \cite{wang2011action}+VLAD \cite{jegou2010aggregating}+ linear SVM}&\textbf{58.1}&63.5&39.0&77.0&72.8&41.7&57.8&36.9&24.2&50.0&54.7&49.0&58.3&38.5&48.9&45.0&84.0&77.4&53.8&67.1&54.9\\
 \hline
 \hline
 \tabincell{c}{Two-stream convnet \cite{simonyan2014two}+Ft+score fusion }&47.6&76.0&40.3&38.9&51.2&26.1&38.0&53.3&21.3&85.2&57.8&73.3&53.4&47.0&28.5&40.7&31.9&51.6&46.8&74.4&49.2\\
  \hline
  \hline
 \tabincell{c}{C3D \cite{tran2014c3d}+linear SVM (FC6)
 }&12.9&59.4&49.0&\textbf{88.3}&79.9&50.0&59.2&49.2&11.6&83.1&27.9&46.1&60.5&19.3&27.4&57.0&85.4&81.6&\textbf{74.7}&74.5&55.2\\
   \tabincell{c}{C3D \cite{tran2014c3d}+Ft+ linear SVM (FC7)
 }&57.8&67.8&38.5&79.3&77.0&49.0&55.9&50.1&26.6&81.6&\textbf{66.0}&36.0&60.9&47.3&53.3&61.3&86.0&78.6&65.6&69.2&60.4\\
  \tabincell{c}{C3D \cite{tran2014c3d}+Ft+softmax }&14.1&50.7&35.6&88.5&70.0&49.2&46.2&59.0&8.4&82.4&39.0&54.1&61.0&16.4&12.4&61.7&89.2&81.8&63.3&74.6&52.9\\
  \hline
  \hline
  \tabincell{c}{DAP3D-Net$^{1}$+Ft (RGB)+softmax}&21.5&69.4&36.4&80.9&75.7&55.4&61.2&66.3&23.6&74.3&53.5&75.1&62.2&41.4&48.7&62.0&87.2&66.8&63.6&70.8&59.8\\
  \tabincell{c}{DAP3D-Net$^{2}$+Ft (AM)+softmax}&26.9&71.7&41.0&73.4&81.1&61.5&67.3&71.9&31.3&79.6&58.1&82.9&67.5&47.4&50.9&66.2&88.6&72.9&66.1&73.5&64.0\\
   \tabincell{c}{DAP3D-Net$^{1}$+BBR+Ft (RGB)+softmax}&25.2&73.4&41.8&81.5&80.0&58.8&64.8&70.2&29.3&78.5&56.7&80.4&65.9&44.0&51.7&65.6&83.4&70.7&67.4&74.9&63.2\\
   \tabincell{c}{\textbf{DAP3D-Net$^{2}$+BBR+Ft (AM)+softmax}}&32.3&\textbf{76.3}&\textbf{53.0}&87.9&\textbf{85.6}&\textbf{66.0}&\textbf{71.8}&\textbf{77.4}&\textbf{34.2}&\textbf{88.1}&61.6&\textbf{87.5}&\textbf{72.8}&\textbf{51.9}&\textbf{57.4}&\textbf{70.7}&91.6&77.4&70.6&\textbf{78.0}&\textbf{69.3}\\
 \hline
\end{tabular}
}
\tiny
\\``Ft" indicates the deep model is fine-tuned with training data from the  HAU dataset. ``Two-stream convnet" is trained with original frames and their optical flows, and score fusion is used for final prediction.  ``RGB" denotes using original data clips to fine-tune our model, while ``AM" denotes using appearance-motion data clips to fine-tune our model. 'soft' indicates directly output the probabilities of each category via C3D deep net instead of using FC6/FC7 feature with linear SVM in \cite{tran2014c3d}. BBR indicates using bounding-box regressor to refine location by the adjustment in Eq. (\ref{eq:e2}). The localization of actions with the methods from 5th to 12th rows in this table is directly using the location ($X_{pro}$,$Y_{pro}$,$T_{pro}$) of the proposals \cite{yu2015fast}.
\vspace{-5ex}
\end{table*}

\begin{figure*}
\vspace{1ex}
  \centering
  \begin{tabular}{cc}
    \includegraphics[width=0.4775\textwidth,height=0.10275\textheight]{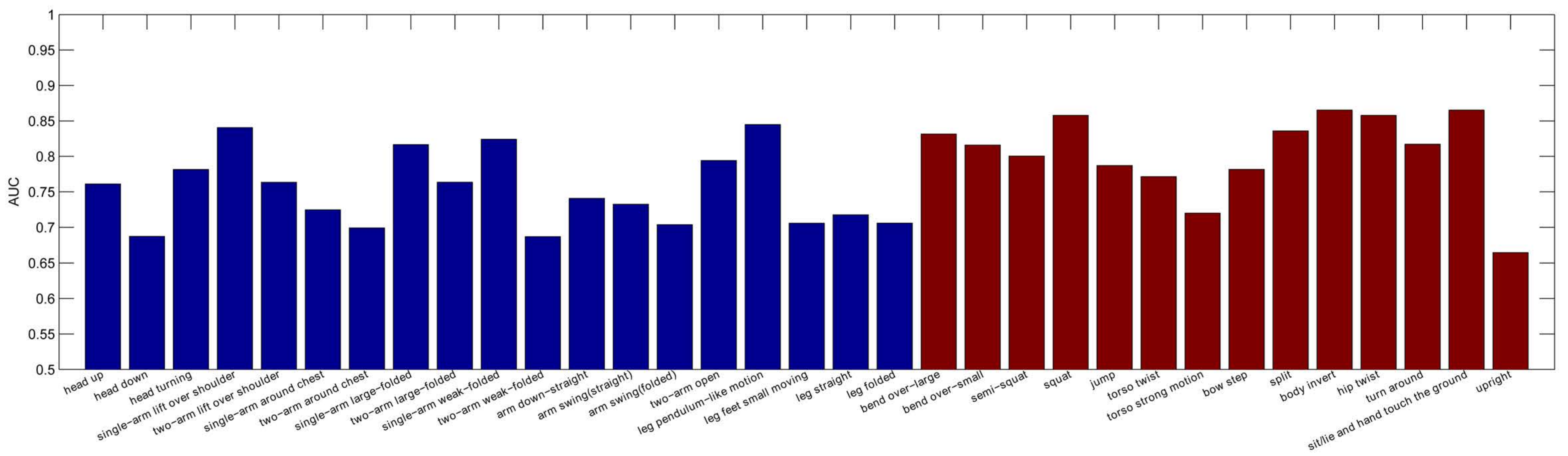} &\includegraphics[width=0.4775\textwidth,height=0.10275\textheight]{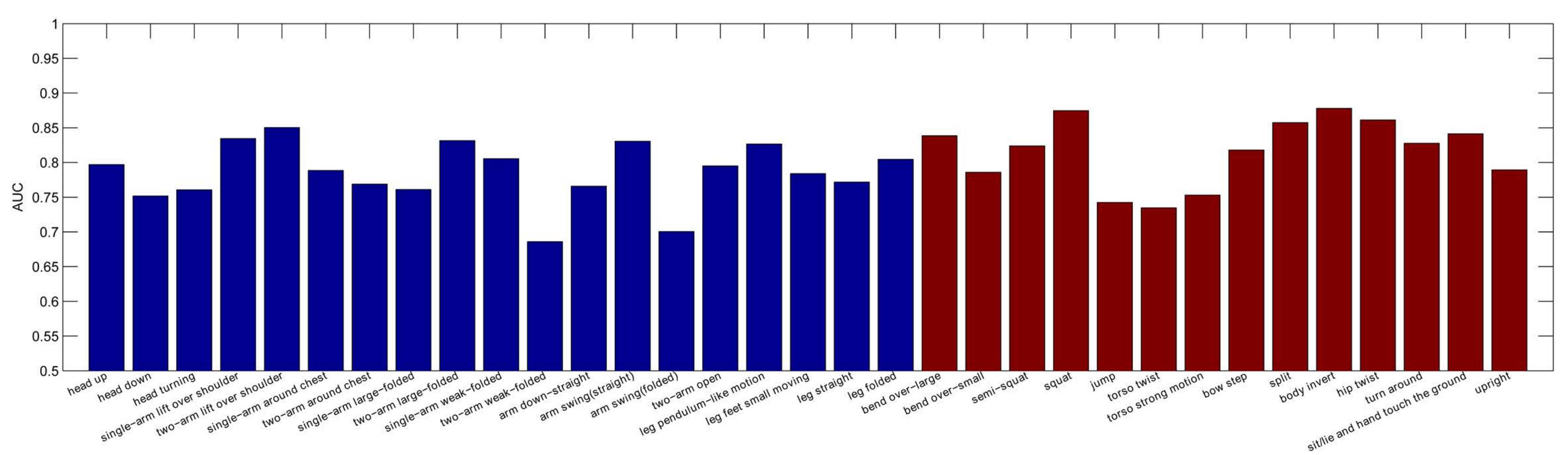}\\
    \footnotesize
 (a) Attributes prediction via DAP3D-Net$^{1}$ (RGB data) & \footnotesize (b) Attributes prediction via DAP3D-Net$^{2}$ (appearance-motion data)\\
  \end{tabular}
  \caption{\footnotesize
  \textbf{AUC of each attribute predicted via DAP3D-Net$^{1}$ and DAP3D-Net$^{2}$ on HAU, respectively}. Blue and red denote the predictions from low-level attributes (H1) and high-level attributes (H2). (Zoom in for better viewing.)}
  \label{attrhau}
  \vspace{-3.5ex}
\end{figure*}

\section{Experiments and results}
\vspace{-0.5ex}
In this section, we evaluate the proposed DAP3D-Net for action parsing in videos on two datasets: NASA and HAU. Our method is implemented using Caffe \cite{jia2014caffe} deep learning framework and we further modify various layers from \cite{tran2014c3d} to specifically fit for our architecture. DAP3D-Net is trained on a workstation configured with GTX TITAN X GPU.

\vspace{-0.05ex}

\subsection{Evaluation on NASA dataset}
The NASA dataset contains $200,000$ aligned action clips with over 300 categories. In our experiments, we only select a subset with the most frequent $100$ action categories, each of which has abundant relevant clips ranging from $540$ to $12900$, as the training set. We further randomly split the subset into training and test sets with a ratio of $9:1$ for evaluating the effectiveness of DAP3D-Net. Following Eq. (\ref{eq:e1}), we train our deep model by stochastic gradient descent ($momentum=0.9$) with the batch size of 40. The initial base learning rate of DAP3D-Net is $\alpha=0.005$, and updated  as $\alpha\rightarrow0.3\alpha$ with the step size of $50K$ iterations. The optimization is terminated at $500K$ iterations. Since each action location in NASA is centrally aligned,  in the testing phase of this experiment, the original test data are directly fed to DAP3D-Net feedforward pass without using action proposal to evaluate action categorization and attributes learning.

To better understand our deep model, two deep nets are trained on NASA with the same architecture of Fig.~\ref{net}: (1) the model trained with original RGB data denoted as DAP3D-Net$^{1}$ and (2) the model trained with previously mentioned appearance-motion data denoted as DAP3D-Net$^{2}$. The numeric values of action categorization accuracies and the mean Area Under the Curve (AUC) of attribute prediction on NASA test set are listed in Table~\ref{table:t1}. Compared with DAP3D-Net$^{1}$ trained on RGB data, DAP3D-Net$^{2}$ can achieve consistently better performance, since appearance-motion contains more information on human motion which is an important factor for action parsing in videos. Fig.~\ref{featuremap} illustrates feature maps from \emph{Conv2} layer of DAP3D-Net$^{2}$, in which salient motion patterns are well learned and shared. We regard DAP3D-Net$^{1}$ and DAP3D-Net$^{2}$ learned from NASA as pre-trained models. In the next sub-section, we will evaluate the effectiveness of our method for multiple-action parsing in realistic videos.

\begin{table}
\vspace{-0.5ex}
\center
\newcommand{\tabincell}[2]{\begin{tabular}{@{}#1@{}}#2\end{tabular}}
\caption{\footnotesize
\textbf{Action categorization accuracy (\%) and attributes prediction AUC on NASA test data.}}
\label{table:t1}
\resizebox{0.45\textwidth}{!}{
\begin{tabular}{|c|c|c|c|}
\hline
\textbf{Method}&\textbf{Accuracy}&\textbf{H1 attributes mean AUC}&\textbf{H2 attributes mean AUC}\\
\hline
DAP3D-Net$^{1}$&76.14&0.711&0.758\\
\hline
DAP3D-Net$^{2}$&\textbf{80.08}&\textbf{0.753}&\textbf{0.792}\\
\hline
\end{tabular}
}
\vspace{-4.5ex}
\end{table}

\begin{figure*}
  \centering
  \includegraphics[width=0.97\textwidth,height=0.242\textheight]{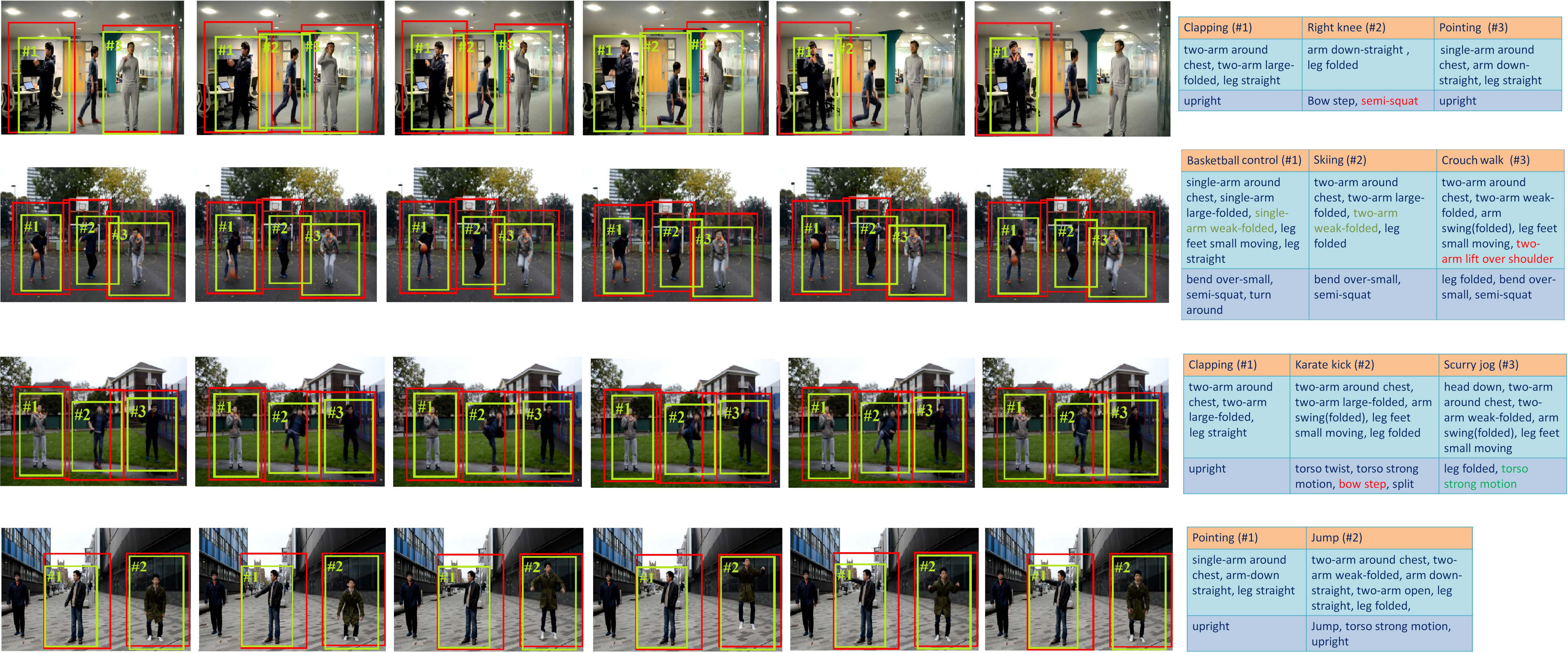}\\
  \caption{\footnotesize
  \textbf{Illustrative example frames for action parsing on the HAU dataset}. The ground-truth location is marked by a red bounding box and the detected bounding box is marked in light green. For attribute prediction, blue indicates corrected prediction, green indicates missed prediction and red indicates wrong prediction. (Better to view this figure in color print.)}
  \label{exampless}
  \vspace{-2.5ex}
\end{figure*}

 \vspace{-0.25ex}
\subsection{Evaluation on HAU dataset}
  \vspace{-0.25ex}
To further evaluate our DAP3D-Net on more challenging and realistic scenarios, we also collect a new Human Action Understanding (HAU) dataset specifically for multiple-action parsing with complex scenes. In particular, the HAU dataset consists of 104 long video sequences, each of which includes several actions performed by different people.  The videos from HAU
contain $C=20$ categories of actions: \begin{small}
\emph{baseballpitch, basketballcontrol, basketballshooting, boilrstok, bowandarrow, clapping, crouchwalk, golfswing, jump, karatekick, pointing, pullrope, rightknee, rounhouse, scurryjog, sitting, skiing, soccer, surfing and walk.}
\end{small}
We annotate the total 1461 aligned actions in all videos from HAU with bounding boxes, action categories and attributes. For this dataset, we randomly select 40 videos as the training set, which contains all action types, to fine-tune our previously learned DAP3D-Net$^{1}$ and DAP3D-Net$^{2}$, and then evaluate the action parsing task on the remaining 64 long videos. In detail, we extract 514 aligned action clips from 40 training videos based on their annotations and then construct the training data following Section ~\ref{rel} to fine-tune DAP3D-Nets with $3.7K$ iterations (The base learning rate is 0.001, and decreases every 1000 iterations by 0.1.). Note that, the number of categories used to fine-tune our DAP3D-Net is $C+1$, since we also add some background clips which are randomly extracted from the non-annotation areas of long training videos as an extra category. In this experiment, we evaluate action parsing on HAU with two aspects: action detection and action attributes learning. Following \cite{yuan2011discriminative,yu2015fast}, for the precision score, a correct action detection is determined if at least $1/8$ of the volume size overlaps a ground-truth. For the recall score, a retrieved ground-truth is determined if at least $1/8$ of its volume size is covered by at least one action detection.

\begin{figure}
  \centering
  \includegraphics[width=0.41\textwidth,height=0.1225\textheight]{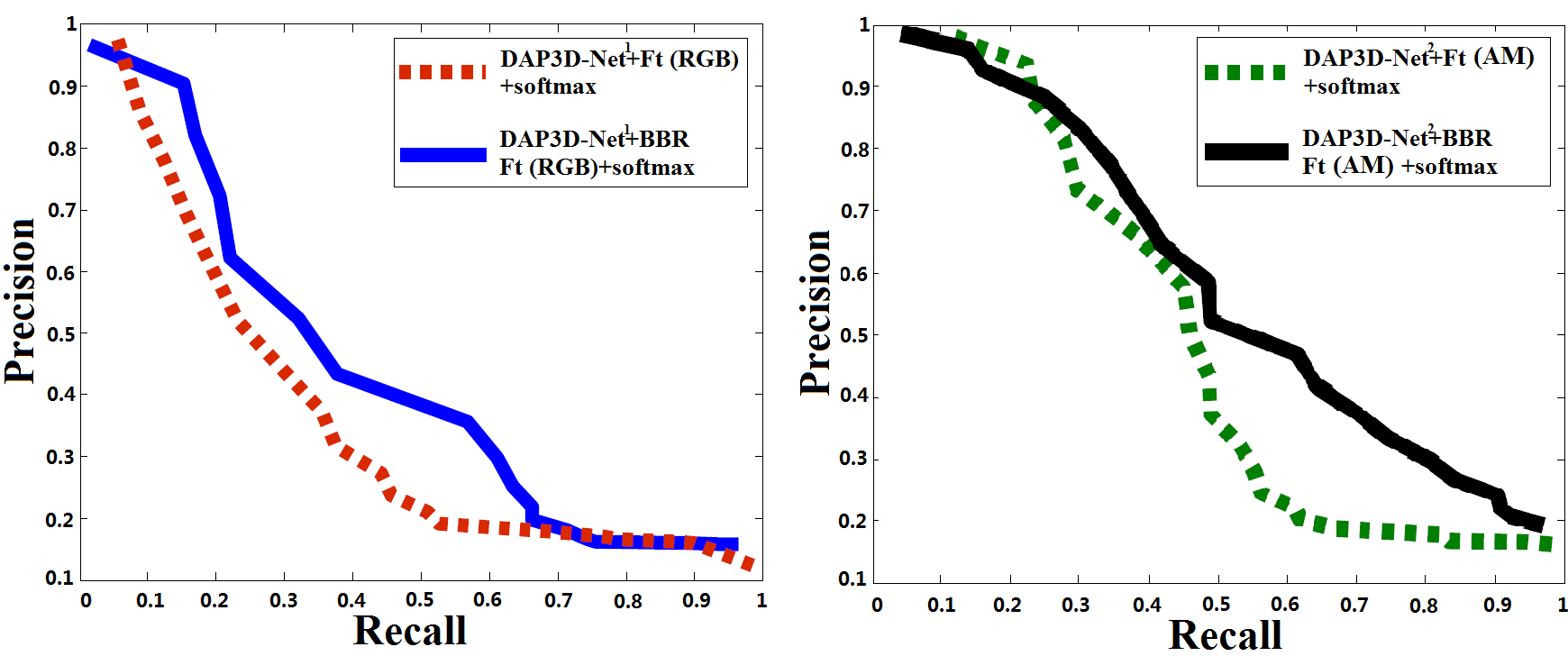}\\
  \caption{\footnotesize
  \textbf{ROC curves for action detection on the HAU dataset}.}
  \label{ROC}
  \vspace{-3ex}
\end{figure}

\vspace{-0.1ex}
The action detection results on HAU are shown in Table~\ref{table:t4}. Specifically, we first compare our method with four state-of-the-art action detection methods: Tubelet \cite{jain2014action}, SDPM \cite{tian2013spatiotemporal}, WSAD \cite{siva2011weakly} and NBMIM \cite{yuan2011discriminative}.
Furthermore, we also extract STIP \cite{laptev2008learning} features and dense trajectory features (DTF) \cite{wang2011action} for both annotated actions from training data and action proposals from test data, respectively. All the features are then embedded into long representations via VLAD \cite{jegou2010aggregating} and fed to SVM. A 2D image-based two-stream deep net \cite{simonyan2014two} pre-trained on UCF101 \cite{soomro2012ucf101} is compared as well. Besides, C3D \cite{tran2014c3d}, as a state-of-the-art 3D deep model pre-trained on the Sport1M dataset, is also used as a feature extractor combining with a SVM classifier for action detection. Instead of extracting features from C3D, the accuracy directly outputted from C3D probability layer with the softmax classifier is also reported. For our method, we fine-tune DAP3D-Net on HAU with RGB and appearance-motion data, respectively. From the results, Tubelet produces the best performance (MAP$=$$55.6\%$) among non-deep-learned methods. While, the STIP+VLAD+linear SVM takes the lowest MAP of 37.0\% in Table~\ref{table:t4} due to the less effective feature extraction. For C3D methods, the results of combining fine-tuned FC7 features with SVM outperform (5.2\%-7.5\% higher) other C3D related works, and the SVM classifier leads to better MAP than directly using softmax via C3D. Our method DAP3D-Net$^{2}$+BBR+Ft+softmax gives the highest MAP of 69.3\% among all the compared methods. The relevant results also illustrate that DAP3D-Net$^{2}$ is superior to DAP3D-Net$^{1}$, which is also reflected with the AUC via precision-recall curves in Fig.~\ref{ROC}. Moreover, bounding box regressor (BBR) in DAP3D-Net can provide a good location adjustment for more accurate (3.4\%$-$5.3\% improvement) action detection compared to DAP3D-Net without using BBR. Besides, Table~\ref{table:t2} shows the results of DAP3D-Net$^{2}$+BBR+Ft+linear SVM on HAU. Combining SVM with extracted FC1 and FC2 features can lead to 1\% improvement over using softmax classifier via DAP3D-Net$^{2}$ but cost much more time for detection, since the separated  pipeline of deep feature extraction and SVM is more time-consuming than directly detecting actions via DAP3D-Net$^{2}$ with softmax classifier (i.e., average 52.34s in total for a 600-frame video).

\begin{table}
\center
\newcommand{\tabincell}[2]{\begin{tabular}{@{}#1@{}}#2\end{tabular}}
\caption{\footnotesize
\textbf{Comparison of attributes prediction results with baselines.} We illustrate our hierarchical learning scheme outperforms learning all attributes from one layer (i.e., FC2) on the HAU dataset.}
\label{table:t3}
\resizebox{0.475\textwidth}{!}{
\begin{tabular}{|c|c|c|c|c|c|c|}
\hline
\textbf{Methods}&\textbf{TrainData}&\textbf{Attribute}&\textbf{\tabincell{c}{Connected\\ Layer}}&\textbf{\tabincell{c}{Mean\\ AUC}}& \textbf{\tabincell{c}{Average \\hit rate}}&\textbf{\tabincell{c}{Std.}}\\
\hline
\hline
&Original&Low-level attributes (H1) &$FC1$&0.772  &16/19&0.071 \\
DAP3D-Net$^{1}$&RGB&High-level attributes (H2) &$FC2$&0.814  &12/14&0.054 \\
+Ft (ours)&data&All attributes (H1+H2) &$FC2$&0.726 &26/33&0.084 \\
\hline
&Appearance-&Low-level attributes (H1) &$FC1$&\textbf{0.795}  &17/19&0.060 \\
DAP3D-Net$^{2}$&motion&High-level attributes (H2) &$FC2$&\textbf{0.847}&12/14&0.042 \\
+Ft (ours)&data&All attributes (H1+H2) &$FC2$&0.759  &28/33&0.075 \\
\hline
\hline
linear SVM&C3D FC6&Low-level attributes (H1)&-&0.678 &14/19 & 0.195\\
linear SVM&C3D FC6&High-level attributes (H2)&-&0.704 &11/14 &0.140\\
\hline
linear SVM&C3D+Ft FC7&Low-level attributes (H1)&-&0.757 &17/19 &0.056 \\
linear SVM&C3D+Ft FC7&High-level attributes (H2)&-&0.783 &11/14 &0.038 \\
\hline
\hline
linear SVM&DTF+VLAD&Low-level attributes (H1)&-&0.602&13/19&0.133 \\
linear SVM&DTF+VLAD&High-level attributes (H2)&-&0.631 &9/14&0.098\\
\hline
\end{tabular}
}
%\tiny
%\\``Ft" indicates the deep model is fine-tuned with training data on HAU datasets
\vspace{-4ex}
\end{table}

After action detection, attributes learning is used for parsing how the action occurs. Table~\ref{table:t3} shows the comparison results on attributes learning between DAP3D-Net and some baseline methods on two-level attributes: H1 and H2. In particular, we extract FC6 features using non-fine-tuned C3D from both HAU ground-truth and proposals, and a linear SVM is then utilized to train independent classifiers with FC6 features on each single attribute. The same procedure is employed  for fine-tuned FC7 features using C3D and DTF+VLAD, as well. In Table~\ref{table:t3}, in general,  multi-output attributes learning via DAP3D-Net produces better performance than independent attributes learning via SVM, and the AUC  calculated on H2 is always higher than that on H1. In detail, DAP3D-Net$^{2}$+Ft significantly outperforms the compared methods with maximum AUC improvements of 32.5\% and 34.2\% for attributes learning on H1 and H2, respectively. Out of 19 attributes on H1 and 14 attributes on H2, it can successfully hit 17 and 12 attributes, respectively. Meanwhile, to evaluate the effectiveness of our two-level hierarchical learning scheme, we also learn all 33 attributes together from the FC2 layer in DAP3D-Net (similar to \cite{shao2015deeply,shankar2015deep}), which proves to perform worse than learning the attributes hierarchically in terms of AUC and hit rates. Fig.~\ref{attrhau} shows the AUC of each attribute prediction via DAP3D-Net$^{1}$ and DAP3D-Net$^{2}$.  Some representative frames of action parsing via DAP3D-Net$^{2}$+Ft+BBR are illustrated in Fig.~\ref{exampless}. It is observed that DAP3D-Net can accurately localize the actions and simultaneously output the action categories and their motion attributes.

\begin{table}
\vspace{-2.75ex}
\center
%\scriptsize
\newcommand{\tabincell}[2]{\begin{tabular}{@{}#1@{}}#2\end{tabular}}
\caption{\footnotesize
\textbf{MAP (\%) of DAP3D-Net$^{2}$+Ft+BBR with softmax vs. linear SVM and the average of total detection time (s) per video on  HAU.}}
\label{table:t2}
\resizebox{0.45\textwidth}{!}{
\begin{tabular}{|c|c|c|c|}
\hline
\textbf{Classifier}&\textbf{DAP3D-Net$^{2}$ (FC2)}&\textbf{DAP3D-Net$^{2}$ (FC1)}&\textbf{DAP3D-Net$^{2}$ (FC2)}\\
\hline
linear SVM&-&70.4 (144.21s)&\textbf{70.6} (158.67s)\\
\hline
softmax&69.3 (\textbf{52.34s})&-&-\\
\hline
\end{tabular}
}
\tiny
\\Total detection time for ``linear SVM" includes action proposal \cite{yu2015fast}, network feedforward pass, deep feature extraction and SVM classification. While, detection time for ``softmax" includes only \cite{yu2015fast} and network feedforward pass.
\vspace{-7ex}
\end{table}

\section{Conclusion}
In this paper, we have developed a multi-task 3D deep convolutional network, i.e., DAP3D-Net, to achieve action parsing by answering where, what and how actions occur in videos, simultaneously. Moreover, two datasets NASA and HAU were contributed for learning and evaluating DAP3D-Net, respectively. Extensive experiments have demonstrated that DAP3D-Net can lead to outstanding performance on action parsing in videos and outperform state-of-the-art methods on action detection and motion attributes learning. In future work, the attributes learned from DAP3D-Net will be further explored with zero-shot learning for recognizing unseen actions.

{\small
\bibliographystyle{ieee}
\bibliography{dap3d}
}

\end{document}